\documentclass{article}
\usepackage{spconf,amsmath,graphicx}

\usepackage{enumitem}
\setlist{nosep, leftmargin=14pt}

\usepackage[dvipsnames]{xcolor}
\usepackage[normalem]{ulem}

\usepackage{mwe} % to get dummy images
\usepackage{multirow}
\usepackage{adjustbox}
\usepackage{makecell}
\title{Preserving Anatomical Continuity: Three-Stage Pipeline for Colon Segmentation in 3D Abdominal CT Scans}
\name{Deshan Kalupahana$^{\star}$, Sonit Singh$^{\star}$, Praveen Ravindran$^{\dagger}$, Arcot Sowmya$^{\star}$}
\address{$^{\star}$School of Computer Science and Engineering, University of New South Wales, Sydney, Australia \\
$^{\dagger}$Sydney Adventist Hospital, Wahroonga, Australia}

\begin{document}
%\ninept
%
\maketitle
%
%===================================================================
%                        ABSTRACT
%===================================================================

\begin{abstract}
Accurate colon segmentation from CT images is essential for colorectal disease analysis, yet deep learning based methods often produce disconnected predictions due to complex anatomy. This study introduces a three-stage, topology-preserving segmentation pipeline to address this issue. The first stage performs initial deep learning-based segmentation, followed by centreline bridging to reconnect disjoint regions and a reconstruction stage to refine continuity. Evaluations on TotalSegmentator and RAOS datasets using overlap, distance and topology-based metrics demonstrate improved structural consistency while maintaining segmentation accuracy. The proposed method enhances topological integrity, enabling more reliable colon segmentation for clinical and research applications.
\end{abstract}
\begin{keywords}
Colorectal Cancer, Colon Segmentation, CT Imaging, Deep Learning, Diverticular Disease, Topology Preservation
\end{keywords}

%===================================================================
%                        INTRODUCTION
%===================================================================
\section{Introduction}
\label{sec:intro}
Structural abnormalities of the colon are key indicators of serious gastrointestinal disorders such as colorectal cancer and diverticular disease. Colorectal cancer is among the most prevalent malignancies worldwide, with approximately 1.9 million new cases and over 900,000 deaths reported in 2022~\cite{zhang_global_2025}. Diverticular disease also represents a major colorectal health concern, affecting up to 70\% of adults over 80 years in Western populations~\cite{strate_epidemiology_2019}. These conditions highlight the need for early and accurate identification of colon abnormalities. Computed tomography (CT) imaging remains the primary modality for detecting such diseases, making colon segmentation a crucial preliminary step in in-depth analysis of the region of interest for accurate identification of these conditions.
%xx

With advancements in deep learning (DL)-based segmentation, automated colon analysis has become increasingly feasible. Early colon segmentation methods relied on thresholding techniques applied to cleansed bowel scans with contrast-enhancing fluids \cite{bert_automatic_2009}. These approaches were later enhanced through DL models capable of addressing partial volume effect \cite{gayathri_devi_automatic_2015}. More recently, several studies have focussed on DL-only segmentation of the colon and sigmoid regions \cite{wang_bowelnet_2023,wu_automatic_2022, rahman_attention_2023}. Wang et al. introduced BowelNet, a two-stage network designed for both colon and small intestine segmentation, achieving an average Dice score of approximately 81\% on a private dataset for colon segmentation~\cite{wang_bowelnet_2023}. The TotalSegmentator framework, trained on 104 anatomical structures including the colon, recorded an 89.6\% accuracy in colon segmentation, yet it also revealed ongoing challenges with discontinuous predictions \cite{wasserthal_totalsegmentator_2023}.

This study aims to resolve the problem of discontinuous predictions by introducing a three-stage, topology-preserving colon segmentation pipeline that ensures structural continuity across disconnected areas while maintaining segmentation accuracy.

%===================================================================
%                        METHODOLOGY
%===================================================================

\section{Method}
\label{sec:methodology}

\begin{figure*}[t]
    \centering
    \includegraphics[width=1\linewidth]{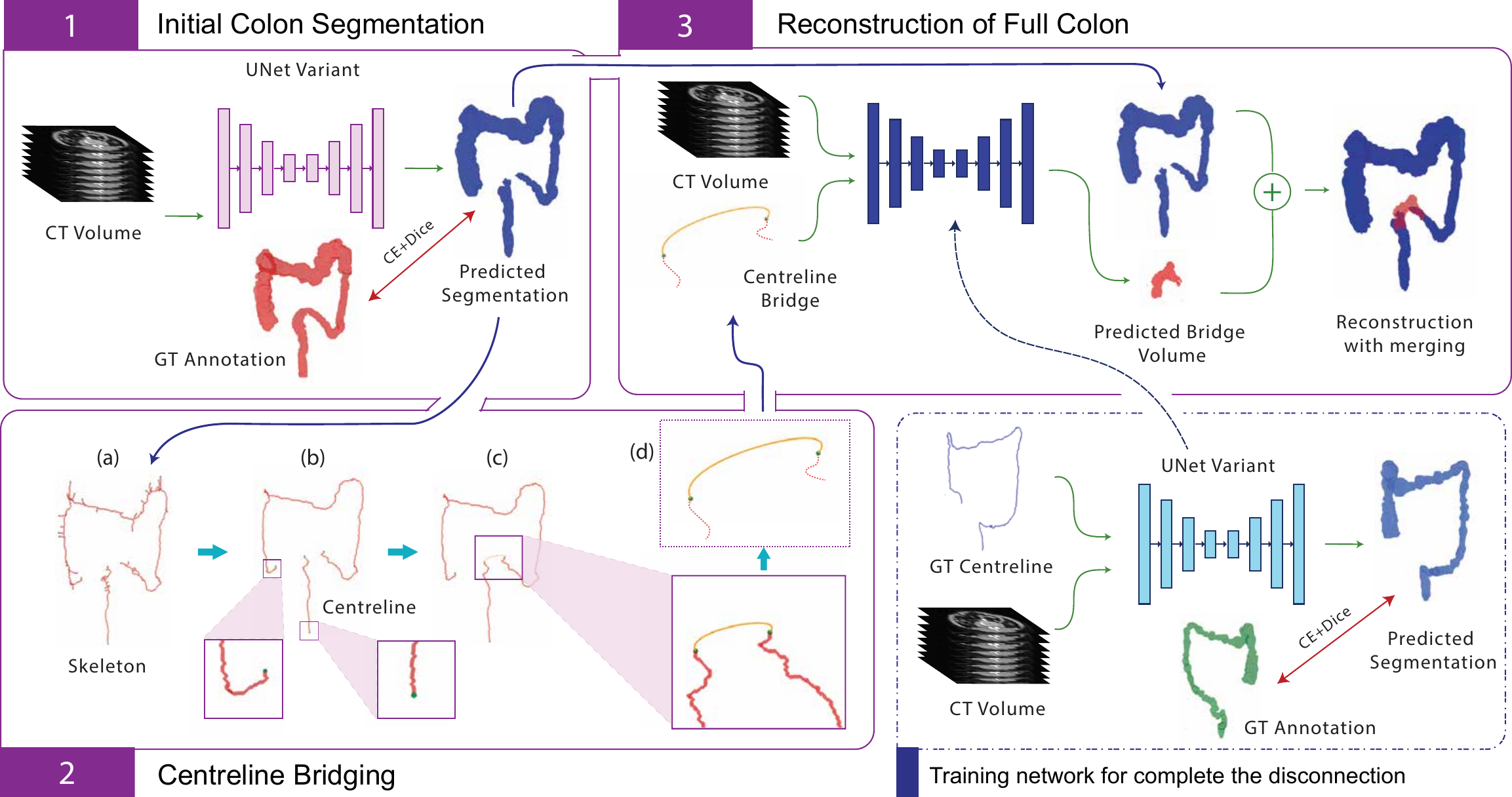}
    \caption{Three-stage pipeline for topology-preserving colon segmentation: (1) initial segmentation with possible disconnections (2) centreline bridging and (3) disconnected region prediction and merging. }
    \label{fig:pipeline}
\end{figure*}

\subsection{Overview of the Segmentation Pipeline}

The proposed topology-preserving colon segmentation pipeline, illustrated in Fig. \ref{fig:pipeline}, comprises three stages. In  Stage 1, the initial colon segmentation is performed, which may contain disconnected regions. Stage 2 constructs a centreline bridge where necessary, based on the extracted centreline from the initial segmentation. Finally, Stage 3 predicts the disconnected regions using the centreline bridge and merges them with the initial segmentation to produce a complete and topologically consistent colon structure.

\subsection{Stage 1 : Initial Colon Segmentation}
This stage focuses on  initial colon segmentation using various segmentation networks. Several UNet-based architectures were assessed, including high-performing medical image segmentation model nnUNet \cite{gao_medical_2025}, the state-of-the-art bowel segmentation model BowelNet \cite{wang_bowelnet_2023}, conventional UNet variants (VNet, UNet++) \cite{gao_medical_2025}, transformer-based models (SwinUNeTR, UNeTR, CoTR) \cite{gao_medical_2025}, and the state-space-based variant U-Mamba \cite{ma_u-mamba_2024}. All models were trained to minimise a combination of cross-entropy (CE) loss and Dice loss between the predicted segmentation and the ground truth (GT) annotations.

\subsection{Stage 2 : Centreline Bridging}
The centreline bridging process, illustrated in Fig. \ref{fig:pipeline}, begins by processing the outputs from Stage 1 to identify connected and disconnected segments. First, a skeleton is generated from the predicted segmentation (Fig. \ref{fig:pipeline} Stage 2 (a)). This skeleton is then converted into a graph representation, where the longest path in each connected component is defined as the main branch, and shorter branches are attached accordingly. To obtain a smooth and topologically consistent centreline, shorter branches, primarily formed due to haustral folds in the colon, are removed. This refinement reduces noise and narrows the potential endpoints for reconnection.

Endpoints are identified as nodes with a single degree in the graph. Colon-specific anatomical knowledge is then applied to differentiate between the cecum and rectum endpoints, which are typically located at the bottom-left and bottom regions of the volume respectively (Fig. \ref{fig:pipeline} Stage 2 (b)). Starting from the component containing the rectum endpoint, the algorithm iteratively searches for the nearest endpoint in disconnected components, progressing toward the cecum endpoint to complete the bridging (Fig. \ref{fig:pipeline} Stage 2 (c)).

This stage does not involve any learning-based operations,  instead it follows a topology-driven approach. Connections between the identified endpoints are created using Bézier curves, constructed according to the directional vectors of the corresponding endpoints (Fig. \ref{fig:pipeline} Stage 2 (c)).

For the next stage, only the centreline bridges are forwarded rather than the entire connected centreline. These bridges include extended points derived from the original centreline, enabling the generation of disconnected regions that remain in contact with the initial segmentation (Fig. \ref{fig:pipeline} Stage 2 (d)).

\subsection{Stage 3 : Reconstruction of Full Colon}
In this stage, a UNet model trained using the ground truth centreline and corresponding CT volume, optimised with CE and Dice losses, was employed to predict the disconnected regions. Since only the centreline bridges are provided as input, the network focusses on reconstructing those specific missing segments. To mitigate potential overpredictions, the predicted components are filtered based on their connectivity to the designated endpoints. Finally, the filtered disconnected regions are merged with the initial segmentation to produce a complete and continuous colon reconstruction.

\subsection{Experimental Setup}
Two CT multi-organ datasets containing colon structures were used in this study. The first dataset is TotalSegmentator (TS) and originally comprised 1,024 samples, from which 467 abdominal scans containing the colon were selected \cite{wasserthal_totalsegmentator_2023}. The second dataset, RAOS, included 413 abdominal CT samples. TS was annotated using an active learning approach, while RAOS was manually annotated \cite{linguraru_rethinking_2024}.

In Stage 1, the best-performing UNet variant for colon segmentation was identified and subsequently used to evaluate the complete three-stage pipeline. For both datasets, the models were trained using five-fold cross-validation with identical data splits to ensure consistency. The outputs from Stage 1 were manually reviewed and grouped to provide a clearer assessment of the pipeline’s performance, enabling analysis of qualitative improvements while ensuring that the initial segmentation accuracy remained unaffected.

Segmentation performance was evaluated using overlap, distance, and topology-based metrics. Overlap metrics included Dice and IoU \cite{muller_towards_2022}, assessing volumetric agreement with the GT annotation. Distance metrics comprised Hausdorff Distance \cite{muller_towards_2022} and Structural Similarity Index (SSIM) \cite{bakurov_structural_2022}, measuring boundary deviation and structural similarity respectively. Topology-based metrics included Centreline Dice (clDice) \cite{shit_cldice_2021}, Average Centreline Distance (ACD) \cite{hutchison_alignment_2012} and Connectivity Ratio (CR), which evaluate continuity, geometric alignment, and structural consistency of the predicted colon segmentation. 
The connectivity ratio is defined as \( \left| 1 - \frac{B_0^{\text{pred}}}{B_0^{\text{GT}}} \right| \), where \( B_0^{\text{pred}} \) and \( B_0^{\text{GT}} \) denote the Betti-0 numbers representing the number of connected components in the prediction and ground truth, respectively. 
Overlap-based metrics are often insensitive to small disconnections that minimally affect region overlap, but can disrupt anatomical correctness. In contrast, topology-based metrics are crucial for evaluating the continuity of the colon and ensuring structurally coherent segmentations.

%xx

%===================================================================
%                        RESULTS AND ANALYSIS
%===================================================================

\section{Results and Analysis}
\label{sec:results}

\subsection{Initial Colon Segmentation}

{\renewcommand{\arraystretch}{1.0}
\begin{table}[]
\caption{Dice scores of Stage 1: Initial Colon Segmentation}
\label{tab:stage_1_results}
\centering

\begin{tabular}{lcc}
\hline
\hline
\textbf{UNet   Variant} & \textbf{TotalSegmentator} & \textbf{RAOS}          \\
\hline
nnUNet                  & \textbf{0.9308±0.0631}    & \textbf{0.8847±0.0706} \\
BowelNet                & 0.9041±0.0848             & 0.8457±0.0929          \\
VNet                    & 0.8882±0.0889             & 0.8348±0.0797          \\
UNet++                  & 0.9118±0.0722             & 0.8594±0.0735          \\
SwinUNeTR               & 0.8701±0.0931             & 0.8513±0.0741          \\
UNetTR                  & 0.8678±0.0892             & 0.8200±0.0803          \\
CoTr                    & 0.9103±0.0802             & 0.8634±0.0886          \\
U-mamba                 & 0.9185±0.0740             & 0.8788±0.0738          \\
\hline
\end{tabular}
\end{table}
}

In Table \ref{tab:stage_1_results}, the results of Stage 1 for both datasets are presented. The nnUNet achieved the best performance for colon segmentation across both datasets, while the other variants performed comparably, with only marginal differences. Qualitative analysis revealed that nearly half of the predicted segmentations closely aligned the ground truth annotations (TS: 49.25\%, RAOS: 52.78\%). However, a subset of the predictions contained disconnected regions (TS: 18.41\%, RAOS: 23.48\%), while in some cases, the predicted results were visually more accurate than the ground truth (TS: 9.20\%, RAOS: 18.40\%). Both situations contributed to a reduction in the Dice score. Disconnections typically occurred due to the loss of fine structural details in the CT volumes. Additionally, the patch-based training approach and the use of Dice loss, which optimises for regional overlap rather than structural continuity, may have limited the network’s ability to preserve the overall colon topology.

% \usepackage{multirow}
{\renewcommand{\arraystretch}{1.2}
\begin{table*}[]
\caption{Experiment results of TS dataset}
\label{tab:ts_stage_3}
\begin{adjustbox}{width=\textwidth}
\begin{tabular}{lcccccccccc}
\hline
\hline
\multicolumn{1}{c}{\multirow{3}{*}{\textbf{Data Category}}} & \multicolumn{2}{c}{\multirow{3}{*}{\textbf{All samples}}} & \multicolumn{8}{c}{\textbf{Samples where   disconnections visible after Stage 1}}                                                                                                                                                                 \\ \cline{4-11}
\multicolumn{1}{c}{}                                        & \multicolumn{2}{c}{}                                      & \multicolumn{2}{c}{\multirow{2}{*}{\textbf{All samples with disconnections after Stage 1}}} & \multicolumn{6}{c}{\textbf{Samples having   continues GT, but disconnections after Stage 1}}                                                        \\ \cline{6-11}
\multicolumn{1}{c}{}                                        & \multicolumn{2}{c}{}                                      & \multicolumn{2}{c}{}                                                                        & \multicolumn{2}{c}{\textbf{All}}                & \multicolumn{2}{c}{\textbf{\textless{}120mm}}   & \multicolumn{2}{c}{\textbf{\textless{}90mm}}    \\ 
\hline
\textbf{No. of Samples}                                     & \multicolumn{2}{c}{467}                                   & \multicolumn{2}{c}{86}                                                                      & \multicolumn{2}{c}{46}                          & \multicolumn{2}{c}{33}                          & \multicolumn{2}{c}{27}                          \\
\hline
\textbf{Metrics}                                            & \textbf{Stage 1}            & \textbf{Stage 3}            & \textbf{Stage 1}                              & \textbf{Stage 3}                            & \textbf{Stage 1}       & \textbf{Stage 3}       & \textbf{Stage 1}       & \textbf{Stage 3}       & \textbf{Stage 1}       & \textbf{Stage 3}       \\
\hline
$\uparrow$ Dice                                                        & \textbf{0.9308±0.0631}      & 0.9285±0.0672               & \textbf{0.8708±0.1077}                        & 0.8619±0.1137                               & \textbf{0.8922±0.0756} & 0.8885±0.0763          & \textbf{0.9145±0.0453} & 0.9126±0.0480          & \textbf{0.9202±0.0471} & 0.9201±0.0494          \\
$\uparrow$ IoU                                                         & \textbf{0.8759±0.0932}      & 0.8726±0.0978               & \textbf{0.7847±0.1443}                        & 0.7720±0.1492                               & \textbf{0.8128±0.1111} & 0.8068±0.1121          & \textbf{0.8455±0.0745} & 0.8426±0.0786          & 0.8554±0.0771          & \textbf{0.8556±0.0803} \\
$\uparrow$ ClDice                                                      & \textbf{0.9551±0.0686}      & 0.9517±0.0740               & \textbf{0.8849±0.1100}                        & 0.8725±0.1185                               & \textbf{0.9028±0.0826} & 0.9010±0.0893          & \textbf{0.9344±0.0498} & 0.9362±0.0546          & 0.9456±0.0459          & \textbf{0.9509±0.0461} \\
$\uparrow$ SSIM                                                        & \textbf{0.9947±0.0039}      & 0.9946±0.0041               & \textbf{0.9929±0.0044}                        & 0.9923±0.0048                               & \textbf{0.9937±0.0038} & 0.9934±0.0042          & \textbf{0.9947±0.0028} & 0.9946±0.0029          & \textbf{0.9948±0.0026} & \textbf{0.9948±0.0026} \\
$\downarrow$ HD95                                                        & \textbf{5.3258±9.3093}      & 5.5621±9.7934               & \textbf{12.8904±15.3848}                      & 13.2122±15.8440                             & 10.0087±8.8628         & \textbf{8.8088±7.9481} & 6.6820±6.1570          & \textbf{6.0453±5.5801} & 5.5017±5.7629          & \textbf{5.0630±5.5834} \\
$\downarrow$ ACD                                                         & \textbf{1.9403±1.8081}      & 1.9940±1.8883               & \textbf{3.5816±2.8222}                        & 3.7196±2.9436                               & 3.0152±1.9857          & \textbf{2.7998±1.8404} & 2.2358±1.1202          & \textbf{2.1148±1.0324} & 2.0948±1.1764          & \textbf{1.9650±1.0523} \\
$\downarrow$ CR                                                          & 0.0974±0.1387               & \textbf{0.0918±0.1536}      & 0.1974±0.2553                                 & \textbf{0.1637±0.3029}                      & 0.1867±0.1460          & \textbf{0.1260±0.1149} & 0.1343±0.1074          & \textbf{0.1090±0.0953} & 0.1254±0.1127          & \textbf{0.1126±0.1033} \\
\hline
\end{tabular}
\end{adjustbox}
\end{table*}
}
% \usepackage{multirow}
{\renewcommand{\arraystretch}{1.2}
\begin{table*}[]
\caption{Experiment results of RAOS dataset }
\label{tab:raos_stage_3}
\begin{adjustbox}{width=\textwidth}
\begin{tabular}{lcccccccccc}
\hline
\hline
\multicolumn{1}{c}{\multirow{3}{*}{\textbf{Data Category}}} & \multicolumn{2}{c}{\multirow{3}{*}{\textbf{All samples}}} & \multicolumn{8}{c}{\textbf{Samples where   disconnections visible after Stage 1}}                                                                                                                                                                      \\ \cline{4-11}
\multicolumn{1}{c}{}                                        & \multicolumn{2}{c}{}                                      & \multicolumn{2}{c}{\multirow{2}{*}{\textbf{All samples with disconnections after Stage 1}}} & \multicolumn{6}{c}{\textbf{Samples having   continues GT, but disconnections after Stage 1}}                                                             \\ \cline{6-11}
\multicolumn{1}{c}{}                                        & \multicolumn{2}{c}{}                                      & \multicolumn{2}{c}{}                                                                        & \multicolumn{2}{c}{\textbf{All}}                  & \multicolumn{2}{c}{\textbf{\textless{}120mm}}     & \multicolumn{2}{c}{\textbf{\textless{}90mm}}     \\
\hline
\textbf{No. of Samples}                                     & \multicolumn{2}{c}{413}                                   & \multicolumn{2}{c}{97}                                                                      & \multicolumn{2}{c}{47}                            & \multicolumn{2}{c}{41}                            & \multicolumn{2}{c}{36}                           \\
\hline
\textbf{Metrics}                                            & \textbf{Stage 1}             & \textbf{Stage 3}           & \textbf{Stage 1}                              & \textbf{Stage 3}                            & \textbf{Stage 1}         & \textbf{Stage 3}       & \textbf{Stage 1}       & \textbf{Stage 3}         & \textbf{Stage 1}       & \textbf{Stage 3}        \\
\hline
$\uparrow$ Dice                                                        & \textbf{0.8847±0.0706}       & 0.8825±0.0717              & \textbf{0.8519±0.0885}                        & 0.8443±0.0885                               & \textbf{0.8634±0.0789}   & 0.8589±0.0781          & \textbf{0.8642±0.0808} & 0.8625±0.0790            & 0.8685±0.0717          & \textbf{0.8689±0.0693}  \\
$\uparrow$ IoU                                                         & \textbf{0.7994±0.1001}       & 0.7962±0.1015              & \textbf{0.7508±0.1165}                        & 0.7393±0.1155                               & \textbf{0.7671±0.1107}   & 0.7599±0.1092          & \textbf{0.7687±0.1132} & 0.7658±0.1106            & 0.7739±0.1040          & \textbf{0.7741±0.1009}  \\
$\uparrow$ ClDice                                                      & \textbf{0.9078±0.1031}       & 0.9017±0.1080              & \textbf{0.8602±0.1241}                        & 0.8389±0.1270                               & \textbf{0.8754±0.1123}   & 0.8566±0.1157          & \textbf{0.8694±0.1162} & 0.8588±0.1216            & \textbf{0.8704±0.1149} & 0.8650±0.1202           \\
$\uparrow$ SSIM                                                        & \textbf{0.9876±0.0056}       & 0.9873±0.0058              & \textbf{0.9864±0.0059}                        & 0.9856±0.0063                               & \textbf{0.9861±0.0057}   & 0.9854±0.0058          & \textbf{0.9859±0.0061} & 0.9854±0.0062            & \textbf{0.9857±0.0062} & 0.9853±0.0063           \\
$\downarrow$ HD95                                                        & \textbf{10.4187±14.5159}     & 10.6874±14.7654            & \textbf{16.2218±18.8779}                      & 17.0241±19.4622                             & \textbf{12.0387±12.4027} & 12.1059±11.9316        & 11.6243±11.6203        & \textbf{11.2389±10.9607} & 10.9384±10.0858        & \textbf{10.3057±9.6813} \\
$\downarrow$ ACD                                                         & \textbf{4.5196±2.6928}       & 4.5763±2.6889              & \textbf{5.5578±3.8153}                        & 5.7576±3.7254                               & \textbf{4.8508±2.5429}   & 4.9573±2.3608          & 4.8731±2.6435          & \textbf{4.8631±2.4374}   & \textbf{4.7193±2.2952} & 4.7378±2.3083           \\
$\downarrow$ CR                                                          & 0.3234±0.1391                & \textbf{0.3082±0.1400}     & 0.3464±0.1650                                 & \textbf{0.2919±0.1668}                      & 0.3653±0.1668            & \textbf{0.3143±0.1695} & 0.3540±0.1736          & \textbf{0.3131±0.1752}   & 0.3447±0.1718          & \textbf{0.3092±0.1785}  \\
\hline
\end{tabular}
\end{adjustbox}
\end{table*}
}

\subsection{Final Colon Reconstruction}

In Tables \ref{tab:ts_stage_3} and \ref{tab:raos_stage_3}, the experimental results obtained after Stage 3 are presented. The evaluation was performed by grouping the samples according to the quality of the Stage 1 predictions. The first category includes all samples (columns 2 and 3). The slight decrease in quantitative evaluation metrics is mainly attributed to cases where the predictions either overpredicted or improved upon the GT annotations. These improvements occur when the pipeline successfully reconstructs missing regions in GT annotations that contain disconnections. 

The second category includes samples that exhibited disconnections in Stage 1 predictions, excluding the best overlapping cases (columns 4 and 5). This category shows a noticeable drop in overlap metric values compared to Category 1. As some of these samples also contain disconnections in the GT, the quantitative improvement is not clearly reflected in the final segmentation results. Specifically, the TS dataset contained 40 ground truth annotations with disconnections, while the RAOS dataset contained 50 such cases.

The third category considers samples where the GT annotations were continuous, but the Stage 1 predictions showed disconnections (columns 6 and 7). This category allows us to quantitatively assess the improvements introduced by the proposed method. The results show notable gains in topology-based metrics, while slight decreases are observed in overlap and distance-based evaluations. These decreases primarily occur in cases with large disconnections where the predicted reconnections do not correspond precisely to the GT annotations. To address this, a distance threshold was introduced, connecting endpoints only when the Euclidean distance between them was less than 120mm (columns 8 and 9) and, in a refined setting, less than 90mm (columns 10 and 11), based on an analysis of all samples. Under these thresholds, improvements are observed across most metrics, especially for disconnections below 90mm.

%TS dataset demonstrated greater improvements compared to the RAOS dataset. This difference can be attributed to the annotation methods used, as RAOS was manually annotated, while TS employed active learning, which tends to produce annotations that align more closely with deep learning–based predictions. 
However, the quantitative improvements remain small, as the reconnected regions are relatively small and therefore contribute only minimally to the overall metric gains.

\subsection{Qualitative analysis}

\begin{figure}[h!]
    \centering
    \includegraphics[width=.75\linewidth]{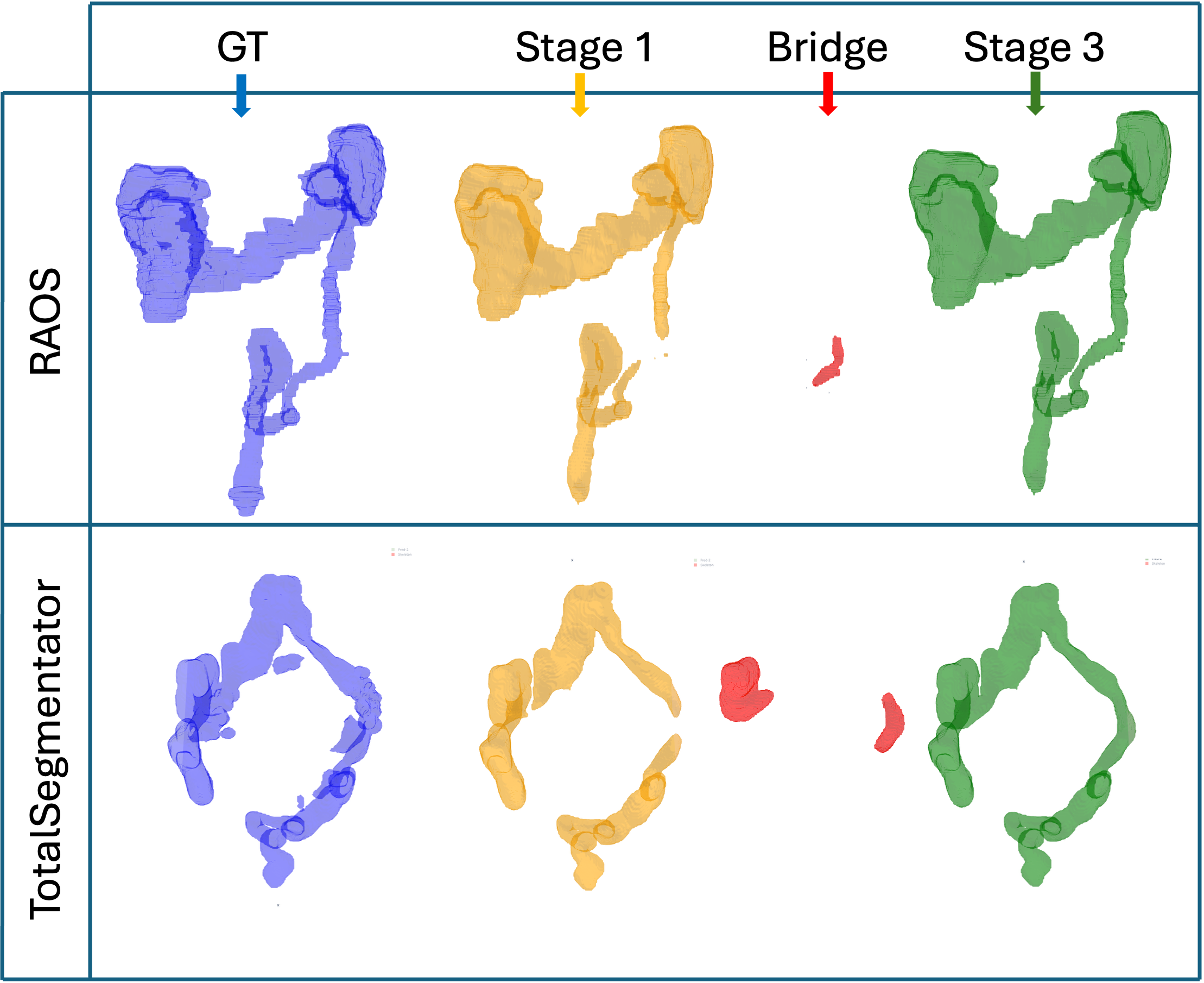}
    \caption{Samples from the TotalSegmentator and RAOS datasets showing cases with disconnections, and the progressive reconstruction of colon segmentation across pipeline stages compared to the ground truth (GT) annotations.}
    \label{fig:segmentations}
\end{figure}

Visual inspection of the results provides important insights into the effectiveness of the proposed method, clearly highlighting the improved continuity and reconnection of colon segments that are not fully reflected by the quantitative metrics. In Fig. \ref{fig:segmentations} sample results from the TS and RAOS datasets after applying the proposed pipeline are shown. The reconnection process performs effectively within the defined disconnection distance threshold for both datasets. However, a few failure cases were observed in the centreline bridging stage. In rare instances, the algorithm fails to accurately identify the cecum and rectum endpoints. This occurs when the colon has sharp bends, resulting in incorrect centreline generation due to faulty skeletonisation. Since the segmentation does not distinguish between the colon wall and lumen, endpoint detection becomes inaccurate. In cases where the colon forms self-connecting loops in segmentation, endpoint identification becomes unreliable. Separately identifying the lumen and colon wall could further enhance the robustness of the proposed method.

%===================================================================
%                        CONCLUSION
%===================================================================

\section{Conclusion}
\label{sec:conclusion}

% This study presented a three-stage, topology-preserving pipeline for colon segmentation from CT images. The proposed framework addresses a common limitation in deep learning-based colon segmentation, where disconnected segments often occur due to complex colon geometry and limited topological awareness in conventional models. By integrating centreline-based bridging and reconstruction, the pipeline effectively restores continuity while maintaining segmentation accuracy. Experimental results on two multi-organ CT datasets, namely TS and RAOS, demonstrate consistent improvements in topology-related metrics, confirming the method’s ability to produce anatomically coherent colon structures. Future work will focus on extending the approach to a learning-based framework that enhances the alignment between predicted and GT centrelines, thereby improving overall segmentation accuracy.

This study presented a three-stage, topology-preserving pipeline for colon segmentation from CT images. The proposed framework addresses a common limitation in deep learning-based colon segmentation, where disconnected segments often occur due to complex colon geometry and limited topological awareness in conventional models. By integrating centreline-based bridging and reconstruction, the pipeline effectively restores continuity while maintaining segmentation accuracy.
%\textcolor{red}{\sout{Experimental results on two multi-organ CT datasets, namely TS and RAOS, demonstrate consistent improvements in topology-related metrics, confirming the method’s ability to produce anatomically coherent colon structures. Future work will focus on extending the approach to a learning-based framework that enhances the alignment between predicted and GT centrelines, thereby improving overall segmentation accuracy.}}
Experimental results on the TS and RAOS datasets show consistent improvements in topology-related metrics, indicating enhanced anatomical coherence of colon segmentations. Although a minor decrease in overlap-based metrics is observed in some instances, future work will evaluate whether this reduction is acceptable by assessing performance on downstream tasks, such as region-of-interest extraction for colorectal cancer and diverticular disease. Further work will also investigate learning-based centreline alignment to enhance segmentation accuracy.


\section{Acknowledgements}
\label{sec:ack}
This research was supported by computational resources provided by the National Computational Infrastructure (NCI), Australia, and Katana, the High Performance Computing facility at the University of New South Wales.
%This work was supported by a UNSW Tuition Fee Scholarship (TFS), plus a stipend for D.K from School of Computer Science and Engineering, University of New South Wales (UNSW). 

\section{Human and/or Animal Subjects}
This study was conducted using publicly available datasets, and therefore did not require ethical approval.

\bibliographystyle{IEEEbib}
\bibliography{content/ref/strings,content/ref/refs}

@article{zhang_global_2025,
	title = {Global burden of colorectal cancer attributable to metabolic risks from 1990 to 2021, with predictions to 2046},
	volume = {25},
	issn = {1471-2407},
	url = {https://bmccancer.biomedcentral.com/articles/10.1186/s12885-025-13643-w},
	doi = {10.1186/s12885-025-13643-w},
	language = {en},
	number = {1},
	urldate = {2025-10-19},
	journal = {BMC Cancer},
	author = {Zhang, Xiaoyue and Xu, Ziqing and Shang, Lin and Yang, Qian and Ye, Hua and Liu, Haiyan and Zou, Yuanlin and Lu, Yin and Zheng, Zhong and Li, Meng and Wang, Peng and Zhu, Jicun},
	month = feb,
	year = {2025},
	pages = {228},
}

@article{strate_epidemiology_2019,
	title = {Epidemiology, {Pathophysiology}, and {Treatment} of {Diverticulitis}},
	volume = {156},
	issn = {00165085},
	url = {https://linkinghub.elsevier.com/retrieve/pii/S0016508519300460},
	doi = {10.1053/j.gastro.2018.12.033},
	language = {en},
	number = {5},
	urldate = {2025-04-16},
	journal = {Gastroenterology},
	author = {Strate, Lisa L. and Morris, Arden M.},
	month = apr,
	year = {2019},
	pages = {1282--1298.e1},
}

@article{gayathri_devi_automatic_2015,
	title = {Automatic {Segmentation} of {Colon} in {3D} {CT} {Images} and {Removal} of {Opacified} {Fluid} {Using} {Cascade} {Feed} {Forward} {Neural} {Network}},
	volume = {2015},
	copyright = {http://creativecommons.org/licenses/by/3.0/},
	issn = {1748-670X, 1748-6718},
	url = {http://www.hindawi.com/journals/cmmm/2015/670739/},
	doi = {10.1155/2015/670739},
	language = {en},
	urldate = {2024-10-20},
	journal = {Computational and Mathematical Methods in Medicine},
	author = {Gayathri Devi, K. and Radhakrishnan, R.},
	year = {2015},
	pages = {1--15},
}

@article{bert_automatic_2009,
	title = {An automatic method for colon segmentation in {CT} colonography},
	volume = {33},
	copyright = {https://www.elsevier.com/tdm/userlicense/1.0/},
	issn = {08956111},
	url = {https://linkinghub.elsevier.com/retrieve/pii/S0895611109000226},
	doi = {10.1016/j.compmedimag.2009.02.004},
	language = {en},
	number = {4},
	urldate = {2024-10-21},
	journal = {Computerized Medical Imaging and Graphics},
	author = {Bert, Alberto and Dmitriev, Ivan and Agliozzo, Silvano and Pietrosemoli, Natalia and Mandelkern, Mark and Gallo, Teresa and Regge, Daniele},
	month = jun,
	year = {2009},
	pages = {325--331},
}

@article{wang_bowelnet_2023,
	title = {{BowelNet}: {Joint} {Semantic}-{Geometric} {Ensemble} {Learning} for {Bowel} {Segmentation} {From} {Both} {Partially} and {Fully} {Labeled} {CT} {Images}},
	volume = {42},
	copyright = {https://ieeexplore.ieee.org/Xplorehelp/downloads/license-information/IEEE.html},
	issn = {0278-0062, 1558-254X},
	shorttitle = {{BowelNet}},
	url = {https://ieeexplore.ieee.org/document/9966840/},
	doi = {10.1109/TMI.2022.3225667},
	language = {en},
	number = {4},
	urldate = {2024-11-03},
	journal = {IEEE Trans. Med. Imaging},
	author = {Wang, Chong and Cui, Zhiming and Yang, Junwei and Han, Miaofei and Carneiro, Gustavo and Shen, Dinggang},
	month = apr,
	year = {2023},
	pages = {1225--1236},
}

@article{wu_automatic_2022,
	title = {Automatic segmentation of colon, small intestine, and duodenum based on scale attention network},
	volume = {49},
	issn = {0094-2405, 2473-4209},
	url = {https://aapm.onlinelibrary.wiley.com/doi/10.1002/mp.15862},
	doi = {10.1002/mp.15862},
	language = {en},
	number = {11},
	urldate = {2024-12-08},
	journal = {Medical Physics},
	author = {Wu, Wenbin and Lei, Runhong and Niu, Kai and Yang, Ruijie and He, Zhiqiang},
	month = nov,
	year = {2022},
	pages = {7316--7326},
}

@article{wasserthal_totalsegmentator_2023,
	title = {{TotalSegmentator}: {Robust} {Segmentation} of 104 {Anatomic} {Structures} in {CT} {Images}},
	volume = {5},
	issn = {2638-6100},
	shorttitle = {{TotalSegmentator}},
	url = {http://pubs.rsna.org/doi/10.1148/ryai.230024},
	doi = {10.1148/ryai.230024},
	language = {en},
	number = {5},
	urldate = {2025-01-26},
	journal = {Radiology: Artificial Intelligence},
	author = {Wasserthal, Jakob and Breit, Hanns-Christian and Meyer, Manfred T. and Pradella, Maurice and Hinck, Daniel and Sauter, Alexander W. and Heye, Tobias and Boll, Daniel T. and Cyriac, Joshy and Yang, Shan and Bach, Michael and Segeroth, Martin},
	month = sep,
	year = {2023},
	pages = {e230024},
}

@misc{ma_u-mamba_2024,
	title = {U-{Mamba}: {Enhancing} {Long}-range {Dependency} for {Biomedical} {Image} {Segmentation}},
	copyright = {Creative Commons Attribution Non Commercial No Derivatives 4.0 International},
	shorttitle = {U-{Mamba}},
	url = {https://arxiv.org/abs/2401.04722},
	doi = {10.48550/ARXIV.2401.04722},
	urldate = {2025-04-19},
	publisher = {arXiv},
	author = {Ma, Jun and Li, Feifei and Wang, Bo},
	year = {2024},
	note = {Version Number: 1},
}

@inproceedings{rahman_attention_2023,
	address = {Port Macquarie, Australia},
	title = {Attention and {Pooling} based {Sigmoid} {Colon} {Segmentation} in {3D} {CT} images},
	copyright = {https://doi.org/10.15223/policy-029},
	isbn = {979-8-3503-8220-4},
	url = {https://ieeexplore.ieee.org/document/10410922/},
	doi = {10.1109/DICTA60407.2023.00050},
	urldate = {2024-12-08},
	booktitle = {2023 {International} {Conference} on {Digital} {Image} {Computing}: {Techniques} and {Applications} ({DICTA})},
	publisher = {IEEE},
	author = {Rahman, Md Akizur and Singh, Sonit and Shanmugalingam, Kuruparan and Iyer, Sankaran and Blair, Alan and Ravindran, Praveen and Sowmya, Arcot},
	month = nov,
	year = {2023},
	pages = {312--319},
}

@article{bakurov_structural_2022,
	title = {Structural similarity index ({SSIM}) revisited: {A} data-driven approach},
	volume = {189},
	issn = {09574174},
	shorttitle = {Structural similarity index ({SSIM}) revisited},
	url = {https://linkinghub.elsevier.com/retrieve/pii/S0957417421014238},
	doi = {10.1016/j.eswa.2021.116087},
	language = {en},
	urldate = {2025-10-27},
	journal = {Expert Systems with Applications},
	author = {Bakurov, Illya and Buzzelli, Marco and Schettini, Raimondo and Castelli, Mauro and Vanneschi, Leonardo},
	month = mar,
	year = {2022},
	pages = {116087},
}

@incollection{hutchison_alignment_2012,
	address = {Berlin, Heidelberg},
	title = {Alignment of {4D} {Coronary} {CTA} with {Monoplane} {X}-ray {Angiography}},
	volume = {7264},
	copyright = {http://www.springer.com/tdm},
	isbn = {978-3-642-32629-5 978-3-642-32630-1},
	url = {http://link.springer.com/10.1007/978-3-642-32630-1_11},
	language = {en},
	urldate = {2025-10-27},
	booktitle = {Augmented {Environments} for {Computer}-{Assisted} {Interventions}},
	publisher = {Springer Berlin Heidelberg},
	author = {Metz, Coert and Schaap, Michiel and Klein, Stefan and Rijnbeek, Peter and Neefjes, Lisan and Mollet, Nico and Schultz, Carl and Serruys, Patrick and Niessen, Wiro and Van Walsum, Theo},
	year = {2012},
	doi = {10.1007/978-3-642-32630-1_11},
	note = {Series Title: Lecture Notes in Computer Science},
	pages = {106--116},
}

@article{muller_towards_2022,
	title = {Towards a guideline for evaluation metrics in medical image segmentation},
	volume = {15},
	issn = {1756-0500},
	url = {https://bmcresnotes.biomedcentral.com/articles/10.1186/s13104-022-06096-y},
	doi = {10.1186/s13104-022-06096-y},
	language = {en},
	number = {1},
	urldate = {2025-10-27},
	journal = {BMC Res Notes},
	author = {Müller, Dominik and Soto-Rey, Iñaki and Kramer, Frank},
	month = dec,
	year = {2022},
	pages = {210},
}

@inproceedings{shit_cldice_2021,
	address = {Nashville, TN, USA},
	title = {{clDice} - a {Novel} {Topology}-{Preserving} {Loss} {Function} for {Tubular} {Structure} {Segmentation}},
	copyright = {https://ieeexplore.ieee.org/Xplorehelp/downloads/license-information/IEEE.html},
	isbn = {978-1-6654-4509-2},
	url = {https://ieeexplore.ieee.org/document/9578225/},
	doi = {10.1109/CVPR46437.2021.01629},
	urldate = {2025-10-27},
	booktitle = {2021 {IEEE}/{CVF} {Conference} on {Computer} {Vision} and {Pattern} {Recognition} ({CVPR})},
	publisher = {IEEE},
	author = {Shit, Suprosanna and Paetzold, Johannes C. and Sekuboyina, Anjany and Ezhov, Ivan and Unger, Alexander and Zhylka, Andrey and Pluim, Josien P. W. and Bauer, Ulrich and Menze, Bjoern H.},
	month = jun,
	year = {2021},
	pages = {16555--16564},
}

@article{gao_medical_2025,
	title = {Medical {Image} {Segmentation}: {A} {Comprehensive} {Review} of {Deep} {Learning}-{Based} {Methods}},
	volume = {11},
	issn = {2379-139X},
	shorttitle = {Medical {Image} {Segmentation}},
	url = {https://www.mdpi.com/2379-139X/11/5/52},
	doi = {10.3390/tomography11050052},
	language = {en},
	number = {5},
	urldate = {2025-10-27},
	journal = {Tomography},
	author = {Gao, Yuxiao and Jiang, Yang and Peng, Yanhong and Yuan, Fujiang and Zhang, Xinyue and Wang, Jianfeng},
	month = apr,
	year = {2025},
	pages = {52},
}

@incollection{linguraru_rethinking_2024,
	address = {Cham},
	title = {Rethinking {Abdominal} {Organ} {Segmentation} ({RAOS}) in the {Clinical} {Scenario}: {A} {Robustness} {Evaluation} {Benchmark} with {Challenging} {Cases}},
	volume = {15009},
	isbn = {978-3-031-72113-7 978-3-031-72114-4},
	shorttitle = {Rethinking {Abdominal} {Organ} {Segmentation} ({RAOS}) in the {Clinical} {Scenario}},
	url = {https://link.springer.com/10.1007/978-3-031-72114-4_51},
	language = {en},
	urldate = {2025-10-27},
	booktitle = {Medical {Image} {Computing} and {Computer} {Assisted} {Intervention} – {MICCAI} 2024},
	publisher = {Springer Nature Switzerland},
	author = {Luo, Xiangde and Li, Zihan and Zhang, Shaoting and Liao, Wenjun and Wang, Guotai},
	year = {2024},
	doi = {10.1007/978-3-031-72114-4_51},
	pages = {531--541},
}

\end{document}